# A Miniaturized Device for Ultrafast On-demand Drug Release based on a Gigahertz Ultrasonic Resonator

*Yangchao Zhou, Moonkwang Jeong, Meng Zhang, Xuexin Duan, Tian Qiu\**

Yangchao Zhou[1,2], Moonkwang Jeong[2], Meng Zhang[3], Xuexin Duan[1], Tian Qiu[3,4,5,*]

[1] State Key Laboratory of Precision Measuring Technology and Instrument, College of Precision Instrument and Opto-electronics Engineering, Tianjin University, Tianjin 300072, China

[2] Cyber Valley Group – Biomedical Microsystems, Institute of Physical Chemistry, University of Stuttgart, Pfaffenwaldring 55, 70569 Stuttgart, Germany

[3] Division of Smart Technologies for Tumor Therapy, German Cancer Research Center (DKFZ) Site Dresden, Blasewitzer Str. 80, 01307 Dresden, Germany

[4] Faculty of Medicine Carl Gustav Carus, Technical University Dresden, Germany

[5] Faculty of Electrical and Computer Engineering, Technical University Dresden, Germany

\*Corresponding authors: Tian Qiu
Email: tian.qiu@dkfz.de

**Abstract**

On-demand controlled drug delivery is essential for the treatment of a wide range of chronic diseases. As the drug is released at the time when required, its efficacy is boosted and the side effects are minimized. However, so far, drug delivery devices often rely on the passive diffusion process for a sustained release, which is slow and uncontrollable. Here, we present a miniaturized microfluidic device for wirelessly controlled ultrafast active drug delivery, driven by an oscillating solid-liquid interface. The oscillation generates acoustic streaming in the drug reservoir, which opens an elastic valve to deliver the drug. High-speed microscopy reveals the fast response of the valve on the order of 1 ms, which is more than three orders of magnitude faster than the start-of-the-art. The amount of the released drug exhibits a linear relationship with the working time and the electric power applied to the ultrasonic resonator. The trigger of the release is wirelessly controlled via a magnetic field, and the system shows stable output in a continuous experiment for two weeks. The integrated system shows great promise as a long-term controlled drug delivery implant for chronic diseases.

## 1. Introduction

Chronic diseases, which are persistent or otherwise long-lasting in their effects or coming with time, such as diabetes, cancer, asthma and so on, are a growing burden of the world.[1,2] They cause long-term suffering to the patients and in some cases may be life-threatening. Traditional drug administration is scheduled with long time intervals, which causes significant decrease of the drug concentration over time and greatly affects the efficacy of the drug. Moreover, high initial concentration may also increase the risk of resistance and cause severer side effects.[3-5] In order to overcome these problems, many efforts have been made to develop drug delivery systems to sustain the drug concentration at the target location.[6,7]

Generally, the drug delivery systems can be categorized as passive and active devices.[8,9] Common passive methods are based on diffusion, which often have a simple structure, for instance, a drug core covered by a polymer membrane.[10,11] And the rate of diffusion is pre-determined by the properties of the coating polymer, which suffers from a decreased drug release rate and is incapable of feedback control.[12,13] Another passive method for drug delivery is osmotic pressure-driven devices.[14,15] They contain an osmotic engine with high-concentration solution to generate hydrostatic pressure, which pushes a movable piston to release the drug from the reservoir. The device does not require an external power supply, and has simple structures and small footprints. However, the release process of these systems is uncontrollable and cannot be stopped once started. Moreover, the rate of drug release decays, as the concentration of the osmotic engine solution drops over time, making them unsuitable for long-term drug delivery.[16,17] Compared to the passive methods, active methods are more controllable. The system converts external power to kinetic energy of the fluid to release the drug from the chamber.[18] For example, thermopneumatic microfluidic pump applies resistive heating to increase the pressure in a sealed chamber to eject drug.[19,20] The method allows for a fast delivery rate of the drug, but it only exhibits slow response to switch on and off, and also consumes a lot of energy. The electromagnetic method uses a drug reservoir covered with a flexible and magnetic membrane. The magnetic membrane is deformed by the applied magnetic field, which allows the drug to be delivered from the chamber through a laser-drilled aperture on the membrane.[21,22] The method cannot stop the diffusion through the aperture at the off-state, when the medication is not needed.[23,24] To summarize, the desired method for drug delivery requires the following characteristics: (1) controlled on-demand drug delivery, (2) stable delivery rate, (3) fast in response, (4) small device volume, (5) low-power consumption and battery-powered.

In recent years, ultrasound has been widely used in medicine for imaging, tissue ablation, contactless manipulation and so on, due to its biocompatibility, high energy efficiency and good spatiotemporal resolution.[25] Common ultrasonic method for drug delivery is based on the acoustic radiation force.[26] Normally, these devices consist of a drug-loaded case and a spring-attached cover that can be moved by the acoustic radiation force, and thus enabling the drug release. However, pushing the mechanical spring to release the drug requires a large acoustic radiation force (~0.1 N), which makes them difficult to miniaturize. Another acoustic method for drug delivery is ultrasound-triggered disruption and self-healing of hydrogels.[27] The ultrasound does not permanently damage the hydrogels but makes it realize temporally short, high-dose bursts of drug exposure. However, this method is not ideal for long-term administration, as the drug slowly leaks from the hydrogel. We recently applied ultrasonic field as a non-contact manipulation method for the actuation of microdevices, for example, to pattern cells,[28] to drive a mini-endoscope,[29] and to perform parallel assembly of microparticles.[30] We also developed a MEMS (Micro-Electro-Mechanical System) gigahertz acoustic device, which has an overall small size on sub-millimeter scale and is able to generate strong acoustic streaming in liquids due to the high attenuation at gigahertz frequency range.[31,32,35,36] The high-speed acoustic streaming has been applied as a micro-manipulation method, for instance, to mix fluids,[37] to dispense droplets,[33] and to manipulate microparticles.[34] However, to the best of our knowledge, it has not been reported that a gigahertz acoustic streaming device can be used for ultrafast on-demand drug delivery.

In this work, we present an acoustofluidic method using a gigahertz resonator to drive high-speed acoustic streaming for on-demand drug delivery. The complete powering, control and actuation system is integrated to a miniaturized capsule with an overall diameter of ~30 mm. Experimental results show that the amount of model drug released can be precisely controlled by modifying the working time and the power applied to the resonator. Besides, with the integrated control module, the trigger of the model drug release can be wirelessly controlled by an external magnet at a distance of 70 mm. We demonstrate that the system has high stability and reliability over a long testing period of two weeks, and it is able to eject liquid model drug on-demand with an ultrafast response time of 1 ms.

## 2. Results and Discussion
### 2.1 The drug delivery device

The drug delivery system consists of a microfluidic module, a power module and a wireless control module. The exploded view of the system is shown in **Figure 1**a. **Figure 1**b shows the detailed 3D schematic of the microfluidic module. Since the essential elements of the device have the characteristic size of micrometer and the device handles nano- to micro-liter fluid, it is named the "microfluidic module". **Figure 1**(c-d) show the cross-sectional view of the section A-A' during off-state and on-state of the resonator, respectively. **Figure 1**e shows the microscopic image of the resonator, where the pentagonal region is the resonant area of the device and the shape was designed to minimize undesired horizontal propagation and improve the energy efficiency. Figure 1f shows the photo of the integrated drug delivery device. The size of the assembled device is 30 mm in diameter, 15 mm in height and 9 g in weight. And the side view the device is shown in Figure S3.

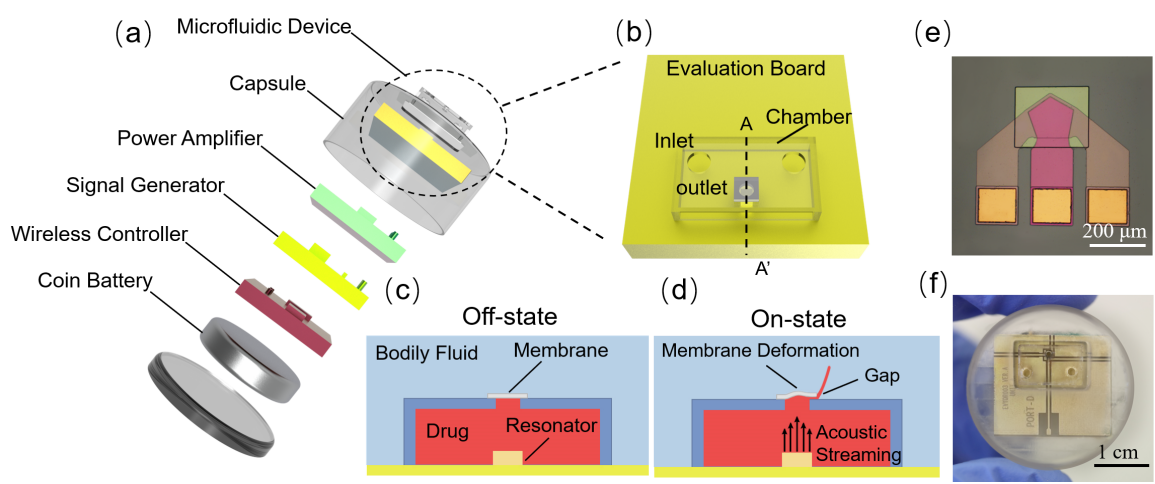

**Figure 1**. On-demand drug delivery system. (a) The exploded view of the drug delivery device. (b) The schematic illustration of the microfluidic module. (c) and (d) The schematic sectional view of section A-A' during off-state and on-state of the resonator, respectively. (e) The top view of the gigahertz acoustic resonator. (f) The assembled drug delivery device.

## 2.2 Theory and simulation of drug delivery process

The resonator utilizes the inverse piezoelectric effect. When an electrical signal at the resonant frequency of 2.56 GHz is applied to the resonator, the transducer converts it into a mechanical vibration, and generates an acoustic wave in the surrounding medium. When the resonator works in a liquid, the acoustic wave attenuates rapidly. Previous studies showed that the attenuation coefficient of ultrasonic waves in water is proportional to the square of the frequency.[33] Thus, the attenuation coefficient of the waves produced by our devices in water to be ~$10^5$ m$^{-1}$, which means that the gigahertz acoustic wave attenuates completely in water within tens of micrometers distance. At the same time, the acoustic energy dissipated in the liquid results in a body force that is perpendicular to the direction of the wave propagation that drives the liquid motion and induces the acoustic streaming.[40] The gigahertz resonator generates a much faster acoustic streaming than low-frequency acoustic devices, thus producing a stronger acoustic streaming force that can, for example, actuating a thin elastic membrane valve in our integrated device.

The schematic of the drug delivery system is shown in **Figure 1**(c-d). When the resonator is at the on-state, the acoustic streaming force that is perpendicular to the membrane pushes the membrane. The deformed membrane leaves a gap on the side, and the drug is released from the gap. As shown in **Figure 2**a, the acoustic streaming force has a linear relationship to the power applied to the device ($R^2$=99.8%). The actuation force applied to the elastic membrane valve can be controlled to actuate the valve's deformation and thus the amount of drug released. **Figure 2**b shows the numerical simulation result of the membrane deformation at an applied power of 1000 mW. The deformation is most pronounced at the center of the membrane for 20 μm and the small gap on the open side also exhibits a maximum deformation of 10 μm in the vertical direction, for the drug to be ejected from the chamber (see also Video S1). Experimental results also demonstrate that the acoustic streaming force is large enough to deform the elastic membrane. **Figure 2**d compares the microscopic images of the cross-section of the membrane before and after applying the acoustic streaming force. The red dashed line represents the membrane position at rest. H represents the displacement of the membrane. **Figure 2**c shows the deformation of the membrane at the gap for different powers applied to the resonator. The maximal deformation increases linearly with the applied power, which is in general agree with the simulation results (Two results show similar growth trends) as shown in **Figure 2**e.

The fluidic flow of the integrated device was also characterized by numerical simulation of the flow fluid. As shown in **Figure 2**f, the red lines represent closed fluidic boundaries and the yellow lines represent open fluidic boundaries. For simplicity, a fixed gap of 10 μm is positioned on the top surface of the reservoir, and 0.5 mm to the left of the resonator, to mimic the real device. 1000 mW of power was applied, and the flow velocity is normalized. Vortices are observed in the drug reservoir, which indicates that the high-speed acoustic streaming mixes the fluid in the chamber and facilitates to homogenize the concentration. Streamlines out of the chamber clearly originate from the gap on the reservoir surface, which indicates that the drug inside the chamber is ejected by the acoustic streaming through the gap, not solely relying on passive diffusion. Similar phenomenon was also observed in the experiments (see Section 2.4). The active streaming device not only enables an ultrafast response (see the Section below) of the device to release the drug by jetting the fluid at the on-state, but also allows a larger elastic restoring force of the elastic membrane to completely seal the chamber at the off-state for long-term on-demand drug delivery without leakage (see Section 2.5).

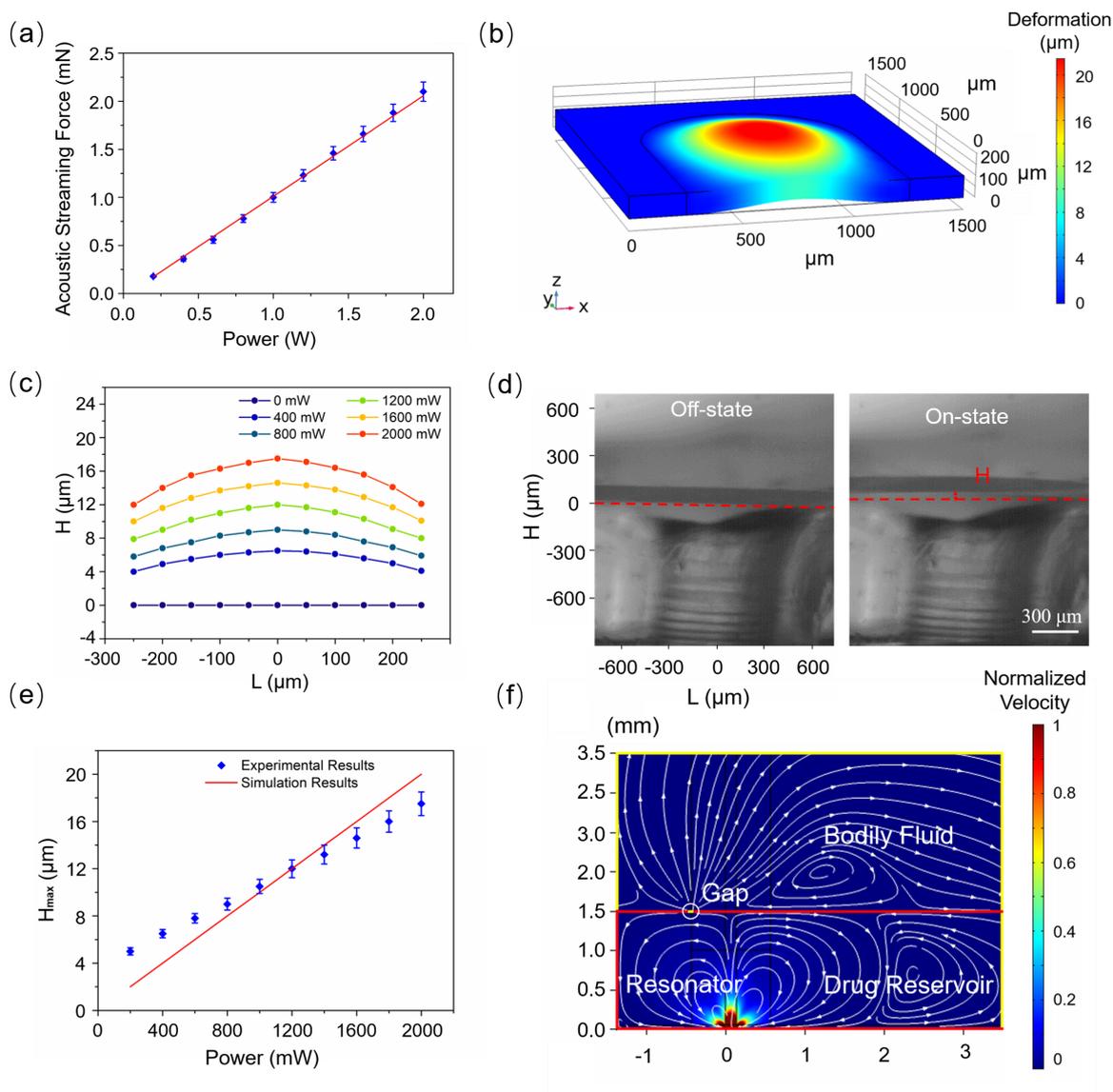

**Figure 2**. Working principle and the simulation of the drug delivery device. (a) The relationship between the applied electric power and the force generated by acoustic streaming. (b) Numerical simulation of the elastic membrane deformation. (c) Measured shapes of the lower edge of the deformed membrane at different electric powers. (d) Microscopic images showing the membrane deformation with the acoustic streaming force. (e) The relationship between the applied electric power and the maximal height of the gap measured in experiments and simulation. (f) Numerical simulation results of the acoustic streaming field in the drug reservoir. Data values represent the average values ± standard deviations.

**2.3 Ultrafast actuation of the elastic valve and rapid drug delivery**

We characterized the time response of the drug ejection process by fluorescent microscopy. To visualize the model drug delivery process, a pulsed signal with a period of 200 ms and a pulse width of 100 ms was first applied. As shown in Video S3 and **Figure 3**a, the size of the fluorescent model drug droplet (visualized as the white area) increases periodically, showing that the fluorescent solution is ejected periodically from the drug reservoir. As the released model drug volume is mostly confined in the 2D space but not forming a 3D droplet, we use the area of the droplet to estimate the amount of released model drug. The released model drug amount is plotted over time in **Figure 3**b, which shows the increase of the amount of released model drug in 1 s selected from the whole drug delivery process as an example. The excitation cycle of 500 mW with a 200 ms total-period and 100 ms pulse-width were repeated for 5 times. After each dosing cycle, an increase of the released model drug amount is observed, which exhibits an average released amount of $0.3 \pm 0.02$ ng with good reproducibility and thus demonstrates the high precision of the drug delivery system. Between the two doses, a plateau is observed in the curve, indicating the complete sealing of the closed valve.

We further decrease the width of the excitation pulse to investigate the lower limit to control the drug release. The speed of switching the elastic valve was tested by measuring the deformation of the membrane in the cross-sectional direction by high-speed microscopy. As shown in **Figure 3**c, a pulsed square-wave signal of 500 mW with a period of 10 ms and a pulse width of 1 ms was applied to the resonator. High-speed video of the membrane deformation is shown in Video S2. **Figure 3**d plots the maximal gap deformation H over time to show the time-response of the elastic valve. The complete cycle of opening and closing of the valve takes ~7 ms. Therefore, in 10 ms the valve is able to operate one full period and returns to the closed state. Ultrafast switching speed is a unique feature of our system. It demonstrates for the first time that gigahertz acoustic streaming can switch an elastic valve at a high speed on the order-of-magnitude of milliseconds, which is very useful for the ultrafast on-demand drug delivery.

When the excitation pulse was decreased to 1 ms (**Figure 3**c), the change of the fluorescent droplet area over one period was no longer visible due to the instrument limitation of high-speed high-resolution microscopy. However, we were able to use fluorescent particle tracking method to verify that there is fluid flow at the valve opening of the device (**Figure 3**e). A pulsed square-wave signal of 500 mW with a period of 500 ms and a pulse width of 1 ms was applied to the resonator. The tracer particles move abruptly right after the resonator is turned off, and keep static while the off-state. **Figure 3**f shows that the average speed of microparticles at the on-state of the device is significantly higher than that of the off-state. These results indicate that the fluidic flow driven by the acoustic streaming can be rapidly switched on and off in 1 ms in accordance to the electric switching of the resonator. The high-speed model drug release experiment is consistent with the measurement of the dynamic response of the membrane deformation, showing the unique capability of the device for ultrafast model drug delivery on the order of milliseconds, which is at least three orders-of-magnitude faster than the state-of-the-art wireless device for drug delivery.[41]

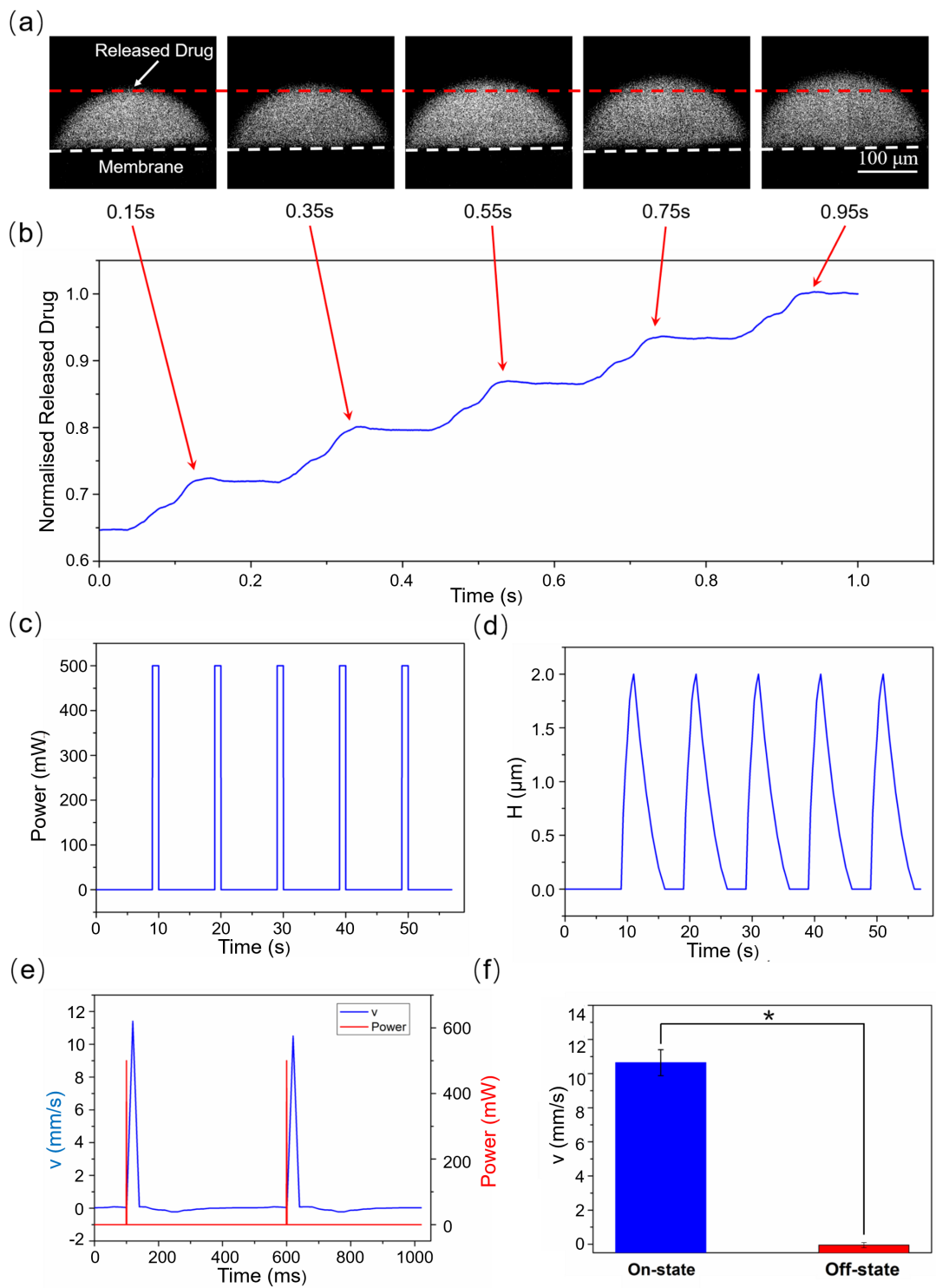

**Figure 3**. Rapid response of drug delivery. (a) Time-lapsed microscopic images showing the model drug release over time. The white area is the model drug droplet released from the gap.

(b) The amount of released model drug plotted over time. 5 cycles with a period of 200 ms and a pulse width of 100 ms were applied. (c) Applied electric signal to trigger the model drug release. (d) The maximal membrane deformation H plotted over time. (e) Tracer particle movement indicates the fluidic flow in response to the excitation pulses of 1 ms width and 500 ms interval. (f) The average speed of tracer particles at the on- and off-states of the device, respectively. Data are presented as the mean ± SD (n = 10), *$p < 0.001$. Data values represent the average values ± standard deviations.

**2.4 The controllability of the drug delivery**

The controllability and the stability are essential for drug delivery systems. In our device, the amount of drug released are mainly determined by the working time and the working power of the acoustic resonator, which can be precisely controlled by the RF signal. The amount of released fluorescence solution was quantified by spectrometry under different electric powers and working time. As shown in **Figure 4**(a-b), the amount of released model drug shows a linear relationship to the applied power from 100 mW to 1000 mW under the same working time of 1 min (with a linear coefficient $R^2$=97.2%). It also shows a linear relationship ($R^2$=99.6%) to the working time from 10 s to 80 s under a constant working power of 500 mW. The results indicate that the amount of released model drug can be precisely controlled in our system by adjusting either the power or the working time of the resonator.

Drug delivery system requires wireless control for in vivo drug delivery. We utilized the sensing of an external magnetic field to control the model drug release. A Hall sensor was integrated in the system to sense the flux density of the magnetic field and to control the RF signal generator. The activation threshold is set to 5 mT, and a permanent magnet is used to regulate the magnetic field around the system by changing the distance between the magnet and the sensor (**Figure 4**c and Fig. S1). When the distance of the magnet is smaller than 70 mm from the sensor, it triggers the electric circuit to power the acoustic streaming. **Figure 4**d shows the release of fluorescent solution under wireless control (see also Video S4). The model drug is ejected from the side of the chamber, when the magnet is close enough (at ~2 s). And when the magnet is moved away (at ~ 32 s), the device shuts off and only diffusion but no active release is observed.

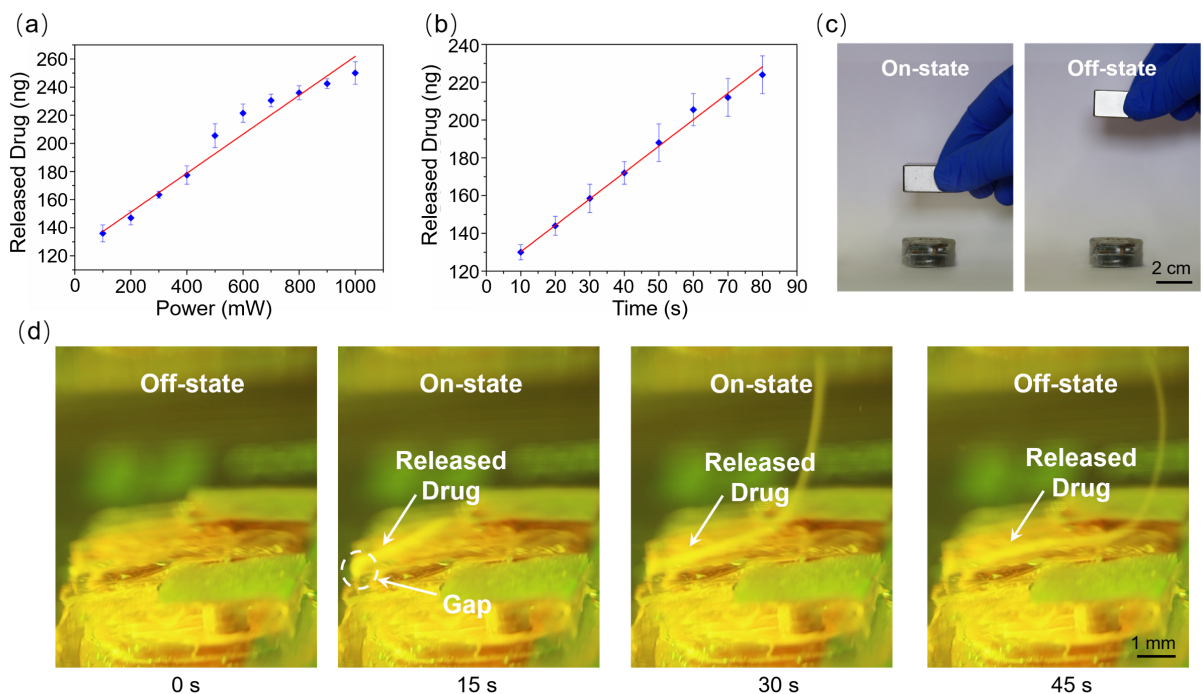

**Figure 4**. Controllable and wirelessly-triggered drug delivery. (a) The relationship between the working power and the amount of released model drug at the working time of 1 min. (b) The relationship between the working time and the amount of released model drug at the working power of 500 mW. Error bars represent standard deviations. (c) Control the switch of the acoustic resonator by adjusting the distance between the magnet and the drug delivery capsule. (d) Snapshots of a video showing the drug release process controlled wirelessly by the magnetic field. Data values represent the average values ± standard deviations.

**2.5 Long-term on-demand drug delivery**

To characterize the on-demand drug delivery capability, a comparison between triggered drug release and diffusion-based drug leakage was performed. As shown in **Figure 5**a, when the device is shut off, the elastic valve keeps a very low leakage rate of $0.12 \pm 0.02$ ng min$^{-1}$, which is negligible comparing to other diffusion-based drug release systems.[8] When the acoustic device is activated to trigger the drug release, for example, for 1 minute at the power of 500 mW in every 100 min, as shown in **Figure 5**b, fast and reproducible model drug release is measured with an average rate of $205.5 \pm 8.5$ ng min$^{-1}$. The model drug release rate exhibits three orders of magnitude increase by the actuation of acoustic streaming. **Figure 5**c illustrates the cumulative amount of model drug released in 600 min. Dramatic increases of the model drug amount can be seen at each on-state period, and the leakage rate remains at a very low level in between two periods of on-states (plateaus in the blue curve), as well as in the negative control experiment while the device is kept at off-state (red curve).

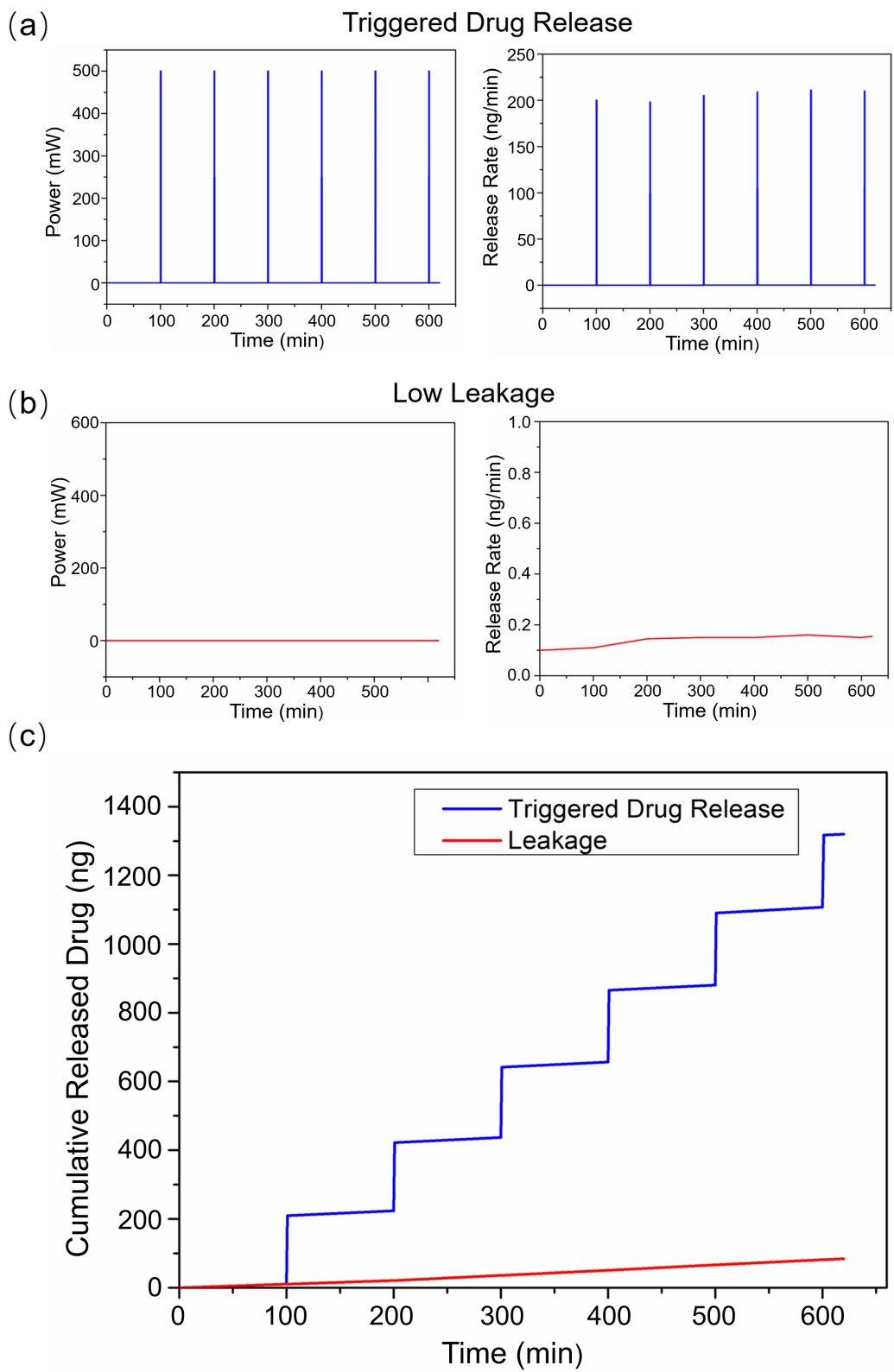

**Figure 5**. Comparison of the drug release based on diffusion and the acoustic streaming triggered release. (a) Without applied electric power, passive-diffusion based drug leakage is very slow and negligible. (b) When the electric power is applied (for 1 min in every 100 min),

synchronized triggered drug release is observed. (c) Cumulative released model drug amount over 600 min of triggered release and passive diffusion, respectively.

We tested the long-term continuous use of the system for two weeks. The model drug was programmed to be released for 1 min each with an interval of 100 min and repeated for 5 cycles per day. The results are shown in **Figure 6**a. During off-state periods, the model drug leakage is negligible as shown in the previous figure. During on-state periods, a steady increase is observed at each release, and the amount is reproducible (shown in the enlarged **Figure 6**b) over 14 days, indicating the long-term stability of our system. The release rate of our system is calculated and shown in **Figure 6**c. Each data point is the average of the released model drug for five times experiments on the same day. The rate of model drug release slightly drops from the initial 220 ± 30 ng min$^{-1}$ to 205 ± 15 ng min$^{-1}$ at Day 5 and is stabilized at this level until the end of the test. The drop in rate is due to the dilution of the drug solution in the drug reservoir, which can be readily compensated by actuating the device for a longer period of time or using a higher electric power for actuation. Meanwhile, the rate of model drug leakage due to passive diffusion remains three orders of magnitude lower at 0.13 ± 0.03 ng min$^{-1}$ over the whole test. The results are consistent with the results for short-term drug delivery in **Figure 5**. The long-term drug delivery experiment is a proof-of-concept showing the stability of our system. Biocompatible encapsulation is essential for future implementation of the device for in vivo drug delivery. Customized housing made of biocompatible materials[42] will be necessary to prevent damage of the device to the host as well as to protect the electronic device from the biological environment. Thanks to the fast release rate of the system, the acoustic resonator only needs be switched on for a short period of time every day. For example, we actuated in total 5 min per day at 500 mW, which consumes the energy of 150 J day$^{-1}$. Using two commercial button cells (e.g., CR2016, each 3 V and 110 mAh), the device can work for around 15 days. At the same time, the ability to wirelessly and precisely control the amount and the time point of drug release offers unique possibility to delivery drug on demand to boost the therapeutic efficacy and lower the side effect.

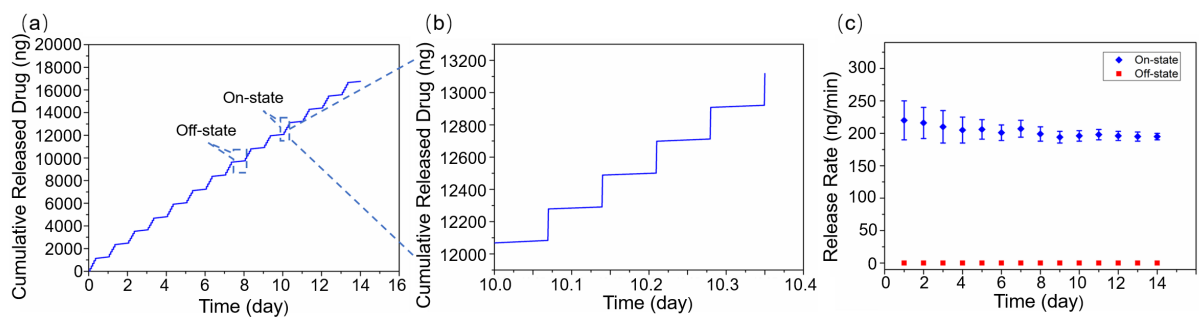

**Figure 6**. Long-term testing of the device for two weeks. (a) Cumulative model drug release by a series of alternative on- and off-states for 14 days. (b) The enlargement of the on-state period per day with five repeated releases of 1 min each and 100 min interval. (c) The average model drug release rates at the on- and off-states of the device, respectively. Data values represent the average values ± standard deviations.

Comparing with the state-of-the-art drug delivery methods (Table 1), the reported device exhibits a much faster response time, a precise control of the drug release rate and a low leakage rate. Many passive devices can only offer a decreasing drug release rate over time, on the contrary, active devices including the one that is developed here exhibit the advantage of a constant release rate. Moreover, since the device is gigahertz RF powered, it has the potential to be further miniaturized in size and suitable for implantation in future.

| Drug delivery device | Response time (s) | Release rate (ng min$^{-1}$) | Leakage rate (ng min$^{-1}$) | Lifetime tested (day) |
|---|---|---|---|---|
| Diffusing device[12] | Uncontrollable | Decreasing | —— | 1 |
| Osmotic devices[15] | Uncontrollable | Decreasing | —— | 3 |
| Thermopneumatic pump[20] | 10 | Decreasing | —— | One-off |
| Magnetic method[22] | 0.06 | Constant (3.4) | 0.053 | 13 |
| Acoustic radiation force[26] | 1 | Constant (6000) | 10 | 2 |
| Self-healing hydrogels[27] | —— | Constant (800) | 80 | 100 |
| This work | 0.001 | Constant (200) | 0.13 | 15 |

**Table 1.** The comparison of different drug delivery devices.

## 3. Conclusion

In this article, we report a miniaturized microfluidic device for wirelessly-controlled ultrafast drug delivery using a gigahertz acoustic resonator. The acoustic energy is converted into kinetic energy of the fluid by the gigahertz-frequency vibration of a solid-liquid interface. The acoustic streaming force pushes a thin membrane to open the elastic valve and release the model drug. The rapid-response of the resonator enables ultrafast model drug delivery on the order of 1 ms. The device can be wirelessly controlled by an external magnetic field. In addition, on-demand model drug delivery was successfully tested for two weeks with negligible leakage at the off-state and instantaneous release at the on-state, demonstrating that the system is suited for long-term controlled drug delivery. The system shows great potential to be used for on-demand implantable drug delivery to treat chronic diseases.

## 4. Experimental Section

*Acoustic resonator fabrication and characterization:* The gigahertz acoustic resonator is fabricated through the standard micro-electro-mechanical system (MEMS) technology. Briefly, Bragg reflectors were formed by depositing multilayers of molybdenum (Mo) and silicon dioxide ($SiO_2$) on the silicon substrate. Then, a layer of aluminium nitride (AlN) was deposited between two layers of Mo on top of the Bragg reflectors to create the acoustic vibration layer. Finally, a layer of $SiO_2$ was deposited as a passivation layer. More detailed information can be found in our previous studies.[38,39]

*Fabrication of the drug delivery device:* The power module consists of a radio frequency (RF) signal generator (HMC386LP4, Analog Devices, USA), a power amplifier (EV1HMC8413LP2F, Analog Devices, USA) and coin batteries. They are connected in series to excite the acoustic resonator. The wireless control module contains a linear magnetic hall sensor (SEN-KY024LM, JOY-IT, Germany) to sense the magnetic field to control the signal generator. The activation threshold of the magnetic sensor was set to 5 mT, based on the characterization of the magnetic field (see Figure S2) of a permanent magnet (Q-30-30-15-N, supermagnete, China) by a 3-axis digital teslameter 3MH3 (3MH3A-0.1%-200 mT, Senis AG, Switzerland). The capsule was designed by CAD design software (Solidworks 2020, Dassault Systemes S.A., USA) and 3D printed on a commercial 3D printer (Formlabs 3L with the Clear Resin material, Formlabs, USA). The resonator was connected to an evaluation board by a wire bonder (ROFS Microsystem, China). A drug chamber (detailed dimensions in Figure S1) made by 3D printing was fixed on the evaluation board. **Figure 1**(c-d) show the cross-sectional view of the section A-A'. A Polydimethylsiloxane (PDMS) membrane (100 μm thick, Wacker Chemie, Germany) was manually cut to an area of ~1.5 mm × 1.5 mm and attached on top of the outlet, with the sides fixed by an epoxy adhesive (Kafuter, China) leaving an opening of 1 mm on the side to the edge of the chamber. As the "model drug", Rhodamine 6G (Sigma-Aldrich, USA) at the concentration of 1 mg mL$^{-1}$ in deionized water, was used as a fluorescent solution to quantify the drug release.

*Quantification of the drug release:* The device was tested in a container with 15 mL deionized water for model drug release, powered by a signal generator (SSG- 6000 RC, Mini-Circuits, USA) and a power amplifier (ZHL-15W-422-S+, Mini-Circuits, USA). The different power from 100 mW to 1000 mW with an interval of 100 mW under the same working time of 1 minute were applied. And the different working time from 10 s to 80 s with an interval of 10 s under a constant power of 500 mW were applied. The concentration of the solution was quantified by a UV-Vis-NIR Spectrophotometer (Cary 6000i, Agilent, USA) and the absorbance peak of Rhodamine 6G at 527 nm was analyzed. The mass of the released model drug was calculated correspondingly. Each experiment was independently repeated for five times.

*Characterization of the acoustic streaming force:* The acoustic streaming force was measured by a commercialized 1-axis force sensor (Force Transducer System 405A, Aurora Scientific, Canada). The probe of the force sensor was placed directly above the acoustic resonator at a distance of 1.5 mm, to measure the acoustic streaming force at the same distance between the resonator and the elastic membrane in the integrated microfluidic device.

*Characterization of the membrane deformation and rapid drug delivery:* The deformation of the membrane was recorded using a high-speed camera (Chronos 2.1 High-Speed Camera, Kron Technologies Inc, Canada) with a macro lens (EF-S 60 mm, Canon, Japan). The release process of Rhodamine 6G solution was recorded by a digital microscope (AM73915MZTL,Dino-Lite, China) at a frame rate of 30 Hz, with the excitation by an LED (CBT-90-G-L11-CM101, Luminus Devices Inc., USA) and an emission filter (Longpass OD4-550nm, Thorlabs, USA).

The fast drug release process was imaged by fluorescent microscopy. To visualize the drug release process at 100 ms, Rhodamine 6G solution was used. A microscope was used, with a LED (M530L4, Thorlabs, USA) as excitation, a filter cube (U-MF2, OLYMPUS, Japan), a camera tube (WA4100, Thorlabs, USA ), a 20× objective (WF5353433422, Mitutoyo, Japan) and a camera (Zyla-5.5, Andor Technology, UK) at a frame rate of 500 Hz. The area of the drug droplet was measured in ImageJ (VERSION, NIH, USA) and correlated to the released volume of the drug. The data was smoothened by a moving mean function with a window length of 11 frames in MATLAB (MathWorks, USA). A period of 1 s was selected from the whole drug release process and shown in Fig. 3b as an example. The overall released dose (calculated as the white area occupied by the droplet) after 1 s was defined as 100% to normalize the curve, and the initial ~65% represents the initial amount of the model drug at the beginning of the 1 s.

To visualize the drug release process at 1 ms, fluorescent particle tracking velocimetry was used. Polystyrene particles (DAFR-T001, CD Bioparticles, China) were used as tracer particles. The same microscope set-up was applied with a recording frame rate at 50 Hz. The moving velocity of particles were calculated for both groups of the on-state and the off-state, respectively. Ten particles were calculated in each group for the statistical analysis. In each experiment, the device was stimulated for at least 5 times, and the experiment was repeated for 5 times. The significance of the difference in fluid velocity between the on- and off-states were analyzed by the student's t-test[27].

*Finite element simulation of the microfluidics:* The acoustic streaming generated by the gigahertz resonator was modeled in previous publications.[33] Here, it was simulated by an oscillating solid-liquid interface at the frequency of 2.56 GHz. Due to the symmetry of the acoustic streaming field, a simplified 2D axial symmetric model was used to simulate the section of A-A' in **Figure 1**b. A body force generated from the acoustic attenuation is applied to an area of 100 μm (width)×50 μm (height) above the resonator. The surrounding material is set to water. The walls of the chamber are set as fixed boundaries, leaving a gap of 10 μm as an open boundary. The flow field was calculated and the flow direction and velocity distribution were plotted.

The deformation of the elastic membrane was simulated using the solid mechanics module in COMSOL Multiphysics 5.5. The membrane is set to the dimension of 1.5 mm (length)×1.5 mm (width)×0.1 mm (height). Three sides of the membrane are set as fixed boundaries, leaving 1 mm on one side as an open boundary. The acoustic streaming force of 1 mN is applied to a circular area with a diameter of 1 mm on the bottom of the membrane. The elastic deformation was calculated and plotted.

**Supporting Information**

Supporting Information is available from the Wiley Online Library or from the author.

**Acknowledgements**

This work was partially supported by the Vector Foundation (RF-robot project, P2020-0111), the MWK-BW for the Cyber Valley Group (Az. 33-7542.2-9-47.10/42/2), the German Cancer Research Center (DKFZ), and the European Union (ERC, VIBEBOT, 101041975). Y. Zhou acknowledges the program of China Scholarships Council (CSC). M. Jeong and T. Qiu acknowledge the support by the Stuttgart Center for Simulation Science (SimTech).

**Conflict of Interest**

The authors declare no conflict of interest.

WILEY-VCH

## The table of contents

*Yangchao Zhou, Moonkwang Jeong, Meng Zhang, Xuexin Duan, Tian Qiu\**

**A Miniaturized Device for Ultrafast On-demand Drug Release based on a Gigahertz Ultrasonic Resonator**

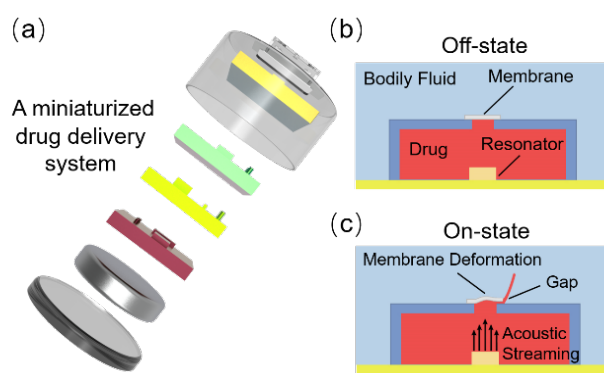

We present an acoustofluidic method using a gigahertz resonator to drive high-speed acoustic streaming for on-demand drug delivery.